\documentclass{article} % For LaTeX2e
\usepackage{iclr2026_conference,times}

\usepackage{amsmath,amsfonts,bm}

\def\eqref#1{equation~\ref{#1}}
\def\1{\bm{1}}

\def\rvepsilon{{\mathbf{\epsilon}}}

\def\rvv{{\mathbf{v}}}

\def\rvx{{\mathbf{x}}}

\def\rmI{{\mathbf{I}}}

\DeclareMathAlphabet{\mathsfit}{\encodingdefault}{\sfdefault}{m}{sl}
\SetMathAlphabet{\mathsfit}{bold}{\encodingdefault}{\sfdefault}{bx}{n}

\def\gL{{\mathcal{L}}}

\def\gN{{\mathcal{N}}}

\newcommand{\E}{\mathbb{E}}

\usepackage[dvipsnames]{xcolor}
\usepackage{hyperref}
\hypersetup{
    colorlinks=true,
 citecolor=MidnightBlue,
 linkcolor=OrangeRed,
 urlcolor=OrangeRed
}

\usepackage{url}
\usepackage{booktabs}       
\usepackage{amsfonts}       
\usepackage{nicefrac}       
\usepackage{microtype}      
\usepackage{color} 
\usepackage{colortbl}
\usepackage{amssymb}
\usepackage{pifont}
\usepackage{tabu}
\usepackage{multirow}
\usepackage{amsmath,bm}
\usepackage{wrapfig}
\usepackage{adjustbox}
\usepackage{enumitem}
\usepackage{cleveref}
\usepackage{subcaption}
\usepackage{graphicx}
\usepackage{arydshln}
\usepackage{afterpage}

\newcommand{\ours}{SVG-T2I}

\definecolor{Gray}{gray}{0.9}

\title{SVG-T2I: Scaling up text-to-image latent diffusion model without variational autoencoder}

\vspace{-15pt}

\author{
\makebox[0.95\textwidth][l]{%
Minglei Shi\textsuperscript{1}\thanks{Equal contribution.} \quad
Haolin Wang\textsuperscript{1}\footnotemark[1] \quad
Borui Zhang\textsuperscript{1} \quad
Wenzhao Zheng\textsuperscript{1} \quad
Bohan Zeng\textsuperscript{2} }\\[0.4em]
\makebox[0.95\textwidth][l]{%
\textbf{~Ziyang Yuan}\textsuperscript{2}\thanks{Corresponding author}\quad
\textbf{Xiaoshi Wu}\textsuperscript{2} \quad
\textbf{Yuanxing Zhang}\textsuperscript{2} \quad
\textbf{Huan Yang}\textsuperscript{2} \quad
\textbf{Xintao Wang}\textsuperscript{2}
} \\[0.4em]
\makebox[0.95\textwidth][l]{%
\textbf{~Pengfei Wan}\textsuperscript{2} \quad
\textbf{Kun Gai}\textsuperscript{2} \quad
\textbf{~Jie Zhou}\textsuperscript{1} \quad
\textbf{Jiwen Lu}\textsuperscript{1}\footnotemark[2]}\\[0.3em]
\makebox[\textwidth][l]{%
\textsuperscript{1}Department of Automation, Tsinghua University \quad
\textsuperscript{2}Kling Team, Kuaishou Technology}
}

\iclrfinalcopy % Uncomment for camera-ready version, but NOT for submission.
\begin{document}

\vspace{-10pt}
\maketitle

\vspace{-26pt}
% ~~~~\textbf{Project Page:} \url{howlin-wang.github.io/svg/} \\
% ~~~~\textbf{Code Repository:} \url{https://github.com/shiml20/SVG}

% \renewcommand{\arraystretch}{1.15} % 可调节比例，1.0 为默认
% \begin{tabular}{@{}ll@{}}
% \textbf{~~~~Project Page:} & \url{https://xxx/SVG-T2I} \\
% \textbf{~~~~Code Repository:} & \url{https://github.com/shiml20/SVG-T2I}
% \end{tabular}
% \renewcommand{\arraystretch}{1.0} % 恢复默认（可选）

\renewcommand{\arraystretch}{1.15}

% 整体右移（可调：0.5em, 1em, 2em）
\hspace{1em}
\begin{tabular}{@{}ll@{}}
\raisebox{-0.2em}{\includegraphics[height=1em]{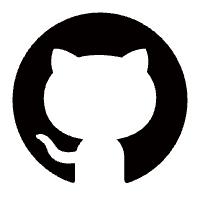}}~\textbf{Code Repository:} &
\url{https://github.com/KlingTeam/SVG-T2I} \\

\raisebox{-0.2em}{\includegraphics[height=1em]{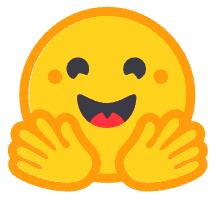}}~\textbf{Model Weights:} &
\url{https://huggingface.co/KlingTeam/SVG-T2I}
\end{tabular}

\renewcommand{\arraystretch}{1.0}

% \vspace{-12pt}
% \vspace{-5pt}

\begin{abstract}
\label{sec:abs}
\vspace{-5pt}

% Visual generation grounded in the visual foundation model (VFM) representations offers a unified pathway for visual understanding, perception, and generation. However, large-scale text-to-image generation trained directly in the VFM representation space has not been thoroughly explored. To address this gap, we scale the SVG (\underline{\textbf{S}}elf-supervised representations for \underline{\textbf{V}}isual \underline{\textbf{G}}eneration) framework to support high-quality text-to-image synthesis within the VFM feature domain. SVG achieves accelerated diffusion training, efficient few-step sampling, and improved generative fidelity. Moreover, experiments demonstrate that SVG preserves the semantic and discriminative strengths of self-supervised representations, offering a principled pathway towards task-general, high-quality visual generation. We fully open-source the project, including the autoencoder and generation model, together with their training, inference, evaluation pipelines, and pre-trained weights, to facilitate further research in representation-driven visual generation.

Visual generation grounded in Visual Foundation Model (VFM) representations offers a highly promising unified pathway for integrating visual understanding, perception, and generation. Despite this potential, training large-scale text-to-image diffusion models entirely within the VFM representation space remains largely unexplored. To bridge this gap, we scale the SVG (\underline{\textbf{S}}elf-supervised representations for \underline{\textbf{V}}isual \underline{\textbf{G}}eneration) framework, proposing SVG-T2I to support high-quality text-to-image synthesis directly in the VFM feature domain. By leveraging a standard text-to-image diffusion pipeline, SVG-T2I achieves competitive performance, reaching 0.75 on GenEval and 85.78 on DPG-Bench. This performance validates the intrinsic representational power of VFMs for generative tasks. We fully open-source the project, including the autoencoder and generation model, together with their training, inference, evaluation pipelines, and pre-trained weights, to facilitate further research in representation-driven visual generation.

\end{abstract}

\vspace{-5pt}
\section{Introduction}
\label{sec:into}

% Generative models, especially text-to-image systems~\citep{ldm,sd3,flux2024}, have achieved remarkable progress in recent years, with diffusion models~\citep{ldm,ddpm,score,lipman2023flowmatching,sd3} emerging as the dominant paradigm. To accelerate the training of VAE-based latent diffusion models (LDM), some methods~\citep{repa, repae, vavae, redi, reg} utilize pretrained VFM features in the way of feature alignment or joint generation. Recent approaches~\citep{svg,rae} have further investigated training diffusion models directly in the high-dimensional VFM feature space, which exhibits superior convergence efficiency in both training and inference, while also paving a promising path toward unifying the diverse vision encoders used across different tasks—for example, CLIP~\citep{clip} and SigLIP~\citep{siglipv2} for visual understanding, VAE~\citep{vae} for visual generation, and VGGT~\citep{vggt} for visual geometry grounding.

Generative modeling, particularly text-to-image synthesis~\citep{ldm,sd3,flux2024}, has advanced rapidly in recent years, with diffusion models~\citep{ldm,ddpm,score,lipman2023flowmatching,sd3} becoming the prevailing solution. To accelerate VAE~\citep{vae}-based latent diffusion models (LDM)~\citep{ldm}, existing works~\citep{repa,repae,vavae,redi,reg} incorporate pretrained Visual Foundation Model (VFM) features through feature alignment or joint generation. 

SVG~\citep{svg} and RAE~\citep{rae} take a further step by training diffusion models directly in the high-dimensional VFM feature space, achieving improved generation quality as well as higher efficiency during both training and inference. In addition to performance gains, this paradigm opens the possibility of unifying the encoder infrastructure across tasks, leveraging a single encoder in place of traditional choices such as SigLIP~\citep{siglip,siglipv2} for understanding, VAE~\citep{vae} for generation, and VGGT~\citep{vggt} for geometry perception and reasoning.

% However, two key questions remain underexplored.
% (1) How can we construct a representation space with strong semantic structure, high reconstruction fidelity, and good alignment with language?
% (2) Are VFM-derived representations suitable for large-scale text-to-image generation, which is critical for real-world applications? 
However, two key challenges remain unresolved:
\begin{enumerate}[leftmargin=*]
    \item Can a unified feature space support visual reconstruction, perception, high-fidelity generation, and semantic understanding without compromising performance on any of these tasks?
    \item Are VFM-derived representations inherently compatible with large-scale, high-resolution text-to-image diffusion training, which is essential for real-world applications?
\end{enumerate}

% Question (1) is important because representations from VFMs inherently suffer from low reconstruction quality. Prior works attempt to address this limitation by either enriching the representation space with additional information~\citep{svg} or modifying the feature space while preserving the underlying semantic structure~\citep{uniflow}. Yet, despite these efforts, there has been no large-scale validation of representation-based generation in VFM feature space, leaving question (2) largely unanswered.
\clearpage
\begin{figure*}[p]
    \centering
    \vspace{10pt}
    \includegraphics[width=\textwidth]{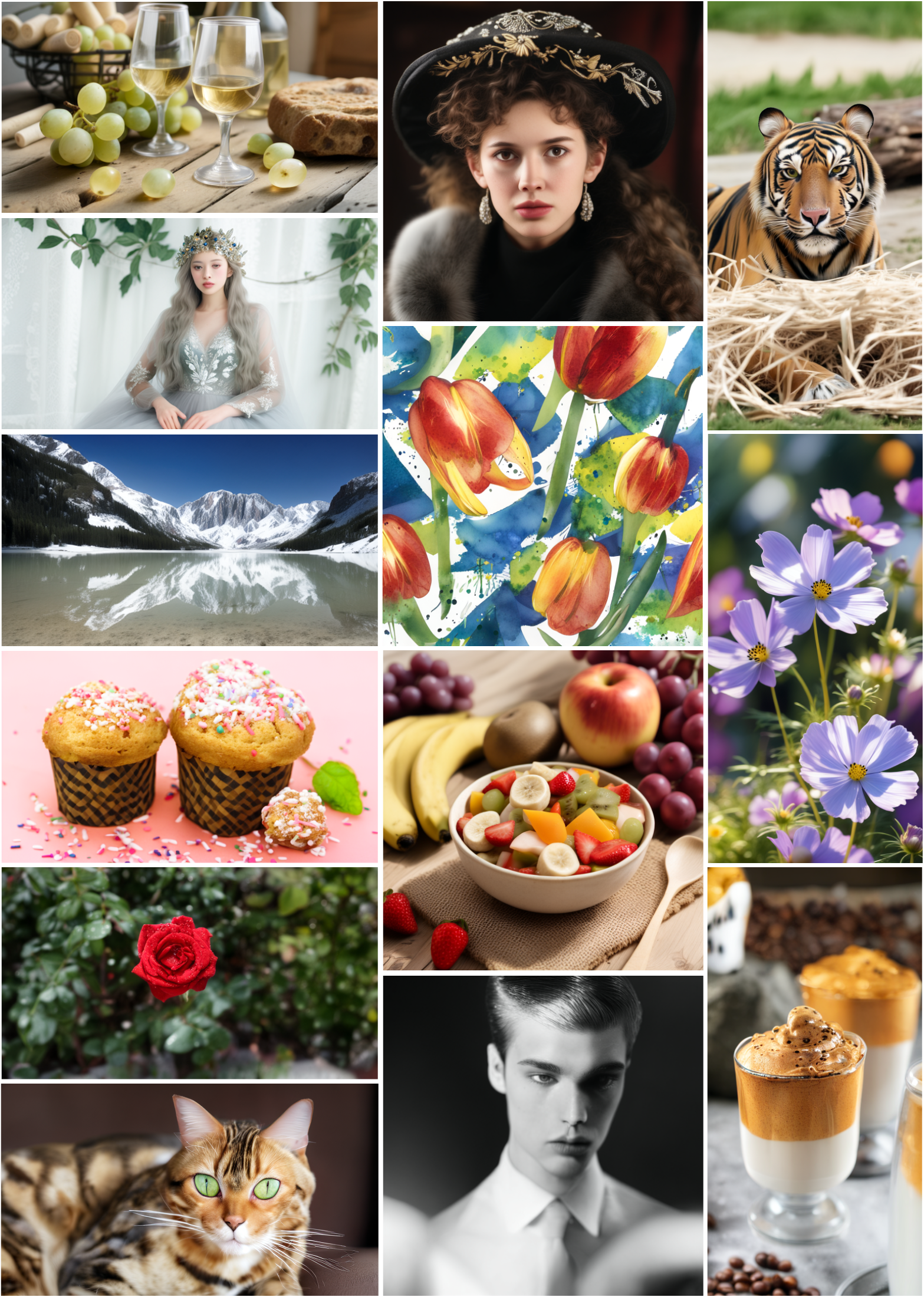}
    \caption{\textbf{Showcases of SVG-T2I with diverse text prompts.} 
    Examples are generated at 720×1280 (left), 1080×1080 (middle), and 1440×720 (right) resolution. All corresponding prompts are listed in~Appendix~\ref{app:prompt}.}
    \label{fig:showcases}
\end{figure*}
\clearpage

Addressing the first challenge is essential for realizing a native, unified general vision model. A principled representation space would eliminate the need for fragmented task-specific encoders, enabling a single architecture to jointly support low-level perceptual fidelity, accurate reconstruction, and high-level semantic understanding and generation. Encouraging progress highlights the feasibility of this goal. Large-scale self-supervised learning~\citep{webssl} has demonstrated that expanding model capacity and training data yields strong perceptual robustness together with competitive performance in understanding tasks. Moreover, frameworks such as SVG~\citep{svg}, UniFlow~\citep{uniflow}, and UniLiP~\citep{unilip} show that VFM representations can be lightly adapted to support precise reconstruction and synthesis while retaining their discriminative semantics. These advances collectively suggest that a unified representation space capable of aligning perception, reconstruction, understanding, and generation is an achievable next step for the field.

Regarding the second challenge, existing approaches~\citep{svg,rae,uniflow} have already demonstrated strong performance on class-conditioned ImageNet~\citep{imagenet} generation. However, large-scale and systematic investigations on text-to-image generation remain largely absent. Given that ImageNet has a limited scale and category diversity, and most vision encoders are already pre-trained on it, such results cannot validate the generalization ability of these methods. As a result, there remain open questions regarding their feasibility and performance potential in realistic text-to-image scenarios.

In this paper, we mainly focus on the second challenge and conduct the first large-scale study on training a text-to-image diffusion model directly within the VFM feature space.

We summarize the primary contributions of this paper as follows:
\begin{itemize}[leftmargin=*]
\item We provide the first large-scale validation of the text-to-image latent diffusion model on the VFM feature space.
\item We open-source the complete training and inference pipeline of the SVG-T2I model, together with pre-trained weights of multiple sizes, to facilitate future research.
\end{itemize}

\section{Related works}
\label{sec:related}

\noindent\textbf{Visual Generative Models.}
Visual generative models aim to approximate the underlying data distribution of real-world visual content and synthesize new samples drawn from it. Over their development, several major paradigms have emerged. 
Generative adversarial networks (GANs)~\citep{gan,wgan,iwgan,stylegan,CycleGAN2017,dcgan,progressivegan,stylegan-xl} generate high-fidelity images via adversarial training but often suffer unstable training, model collapse, and poor interpretability. Autoregressive models~\citep{pixelcnn,imagetransformer,imagegpt,llamagen,VAR} frame generation as a sequential prediction task, achieving strong distribution modeling and stability through likelihood maximization, yet they incur high inference latency due to their token-by-token nature. Masked generative models~\citep{maskgit,maskdit,magvit-v2,mar} predict missing tokens given visible context, analogous to masked language models in NLP, also achieving strong performance. Recently, Diffusion models~\citep{ddpm,iddpm,ddim,score} have been established as the leading paradigm by learning to invert a progressive noising process. They offer robust optimization objectives, superior mode coverage. An enhanced approach, the latent diffusion model (LDM)~\citep{ldm,dit,sit,rectifiedflow}, incorporates a VAE~\citep{vae} into the diffusion pipeline to operate within a lower-dimensional latent space, thereby reducing computational cost while preserving generation quality. However, the latent space of VAEs lacks coherent semantic structure, making it suboptimal for downstream understanding and perception tasks.

% \noindent\textbf{Visual representation learning.}

\noindent\textbf{Representation for Generative Modeling.}
Recent works have increasingly focused on improving the representation of generative modeling. One direction enhances the VAE latent space through structural regularization~\citep{betavae,spherevae,eqvae,diffusability} or diffusion-guided reconstruction~\citep{diffusionautoencoder,diffusevae,sgm}. Another line aligns generative latents with external semantic embeddings~\citep{repa,repae,vavae,rcg} or incorporates discriminative knowledge in generation~\citep{rcg,reg,redi}. More recently, SVG~\citep{svg} and RAE~\citep{rae} demonstrate strong performance by training diffusion models directly in the VFM feature space. Concurrent works such as UniLiP~\citep{unilip} and UniFlow~\citep{uniflow} demonstrate that self-distilled VFMs, when coupled with a pixel decoder and trained under MAR~\citep{mar} or MetaQueries-style paradigms~\citep{metaqueries}, can effectively support both high-quality reconstruction and generation.

\section{Methodology}
\label{sec:method}

\begin{figure}[t]

    \centering
    \vspace{-25pt}    \includegraphics[width=\linewidth]{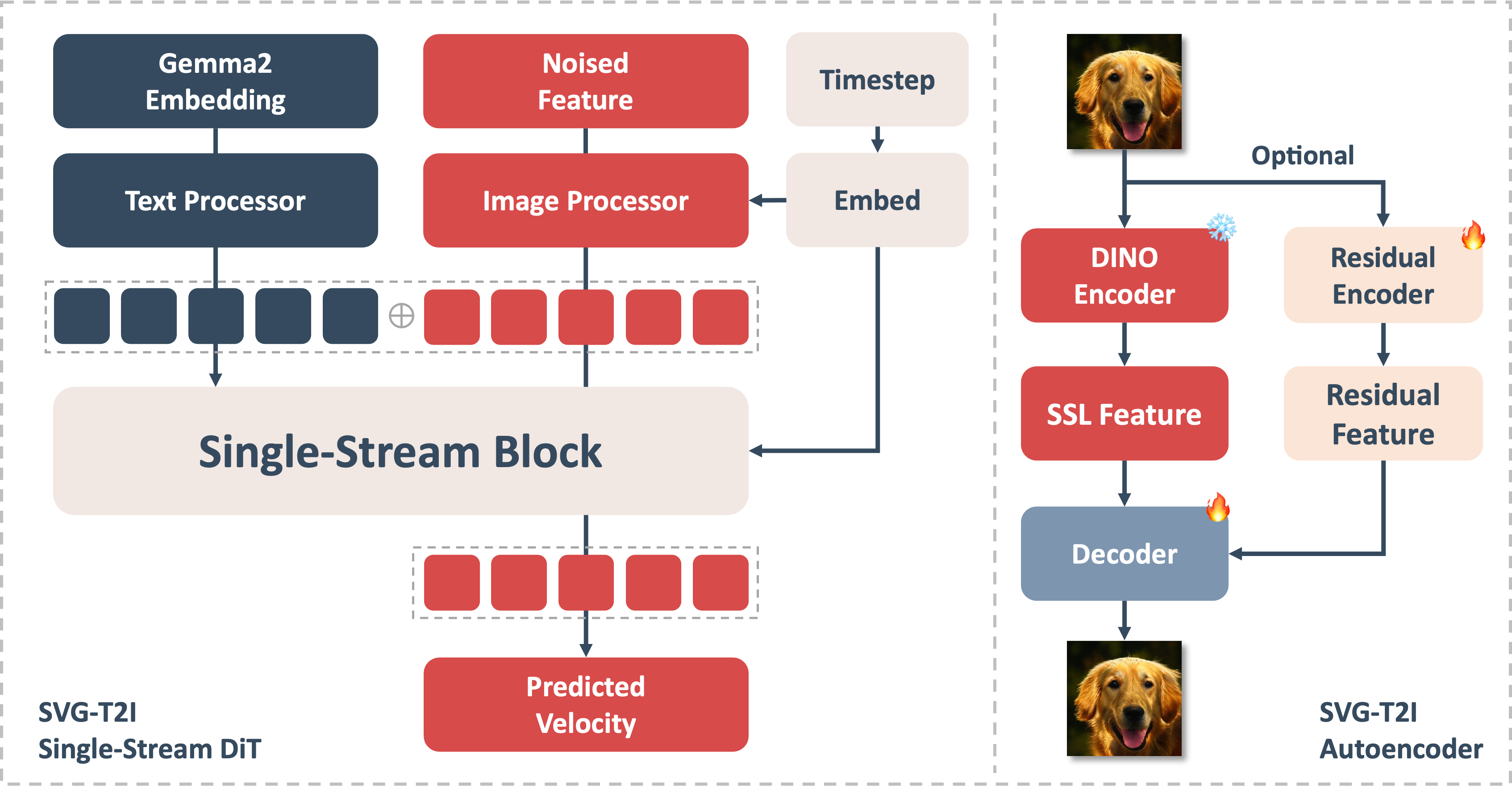}

    \caption{\textbf{Architecture of the proposed~\ours. } (Left) Single-Stream DiT architecture. (Right) We optionally augment the DINO encoder with a Residual Encoder to achieve high-quality reconstruction and preserve transferability.}
    \label{fig:architecture}
    % \vspace{-13pt}
\end{figure}

%We keep the DINO encoder fixed, and only the Residual Encoder and decoders are trainable.
\subsection{Preliminaries}
\noindent\textbf{Diffusion models.}
Diffusion Models~\citep{ddpm,ldm,score} have been the dominant generative modeling for continuous feature space, which can transform the Gaussian distribution to the data distribution through iterative inference. In this paper, we adopt the widely used flow matching method~\citep{rectifiedflow, lipman2023flowmatching, sd3} for training~\ours. Flow matching constructs a velocity field that interpolates between a Gaussian distribution and the data distribution:
\begin{align}
    \rvx_t &= (1-t)\rvx_0 + t\rvepsilon, 
    \quad t \in [0,1], \quad \rvepsilon \sim \gN(0, \rmI),
    % \rvv_t &\triangleq \frac{\mathrm{d} \rvx_t}{\mathrm{d} t} 
           % = \rvepsilon - \rvx_0 .
\end{align}
The flow matching objective is then formulated as
\begin{equation}\label{equ:fm}
    \gL_{\mathrm{FM}} = \E_{\rvx_0\sim p_0(\rvx),\rvepsilon \sim p_1(\rvx)} [\lambda(t)\|\rvv_\theta(\rvx_t, t)-(\rvepsilon - \rvx_0)\|] .
\end{equation}
Sampling from a flow-based model can be achieved by solving the probability flow ODE.

\subsection{Self-supervised Representations for Visual Generation}
SVG~\citep{svg} demonstrated the feasibility of achieving high-quality image reconstruction and class-to-image generation within high-dimensional VFM feature spaces. Building on this foundation, SVG-T2I extends the approach to large-scale text-to-image training, enabling effective generation directly in the VFM feature domain. The overall architecture of~\ours~is shown in \Cref{fig:architecture}.

% \noindent\textbf{\ours~Autoencoder.} Following the practice of SVG~\citep{svg}, RAE~\citep{rae}. We release two kinds of autoencoders to boost the research of the community. The first variant, denoted as \textit{autoencoder-P (Pure)}, utilizes frozen DINO features directly. The second variant, denoted as \textit{autoencoder-R (Residual)}, extends the frozen DINO features with a residual compensatory branch built on a Vision Transformer~\citep{vit}, designed to enhance high-frequency details and mitigate color cast artifacts. All autoencoder variants utilize the same decoder following the decoder design from ~\citep{svg}, which maps the feature back to pixel space. 

\noindent\textbf{\ours~Autoencoder.} Inheriting the architectural design from SVG~\citep{svg} and RAE~\citep{rae}, we release two autoencoder configurations to facilitate community research. The first, \textit{autoencoder-P (Pure)}, utilizes frozen DINOv3 features directly. The second, \textit{autoencoder-R (Residual)}, retains the residual branch design from SVG as an optional choice. This residual module, built on a Vision Transformer~\citep{vit}, is designed to compensate for high-frequency details and color cast artifacts when higher fidelity is required. Both variants utilize the same decoder design to map features back to pixel space.

% The third variant, denoted as \textit{autoencoder-D (Distill)} unfrozen the parameters of DINO encoder with distill loss \textcolor{red}{TODO} to enhance the reconstruction capability while maintain the semantic property.

\begin{figure*}[t]  % [t] 表示优先放在页顶；双栏跨栏用 figure*，单栏用 figure
    \centering  % 图片居中对齐
    % 引入图片：width=0.8\textwidth 控制宽度（80%文本宽，避免过满），height可省略（按比例缩放）
    \includegraphics[width=\textwidth]{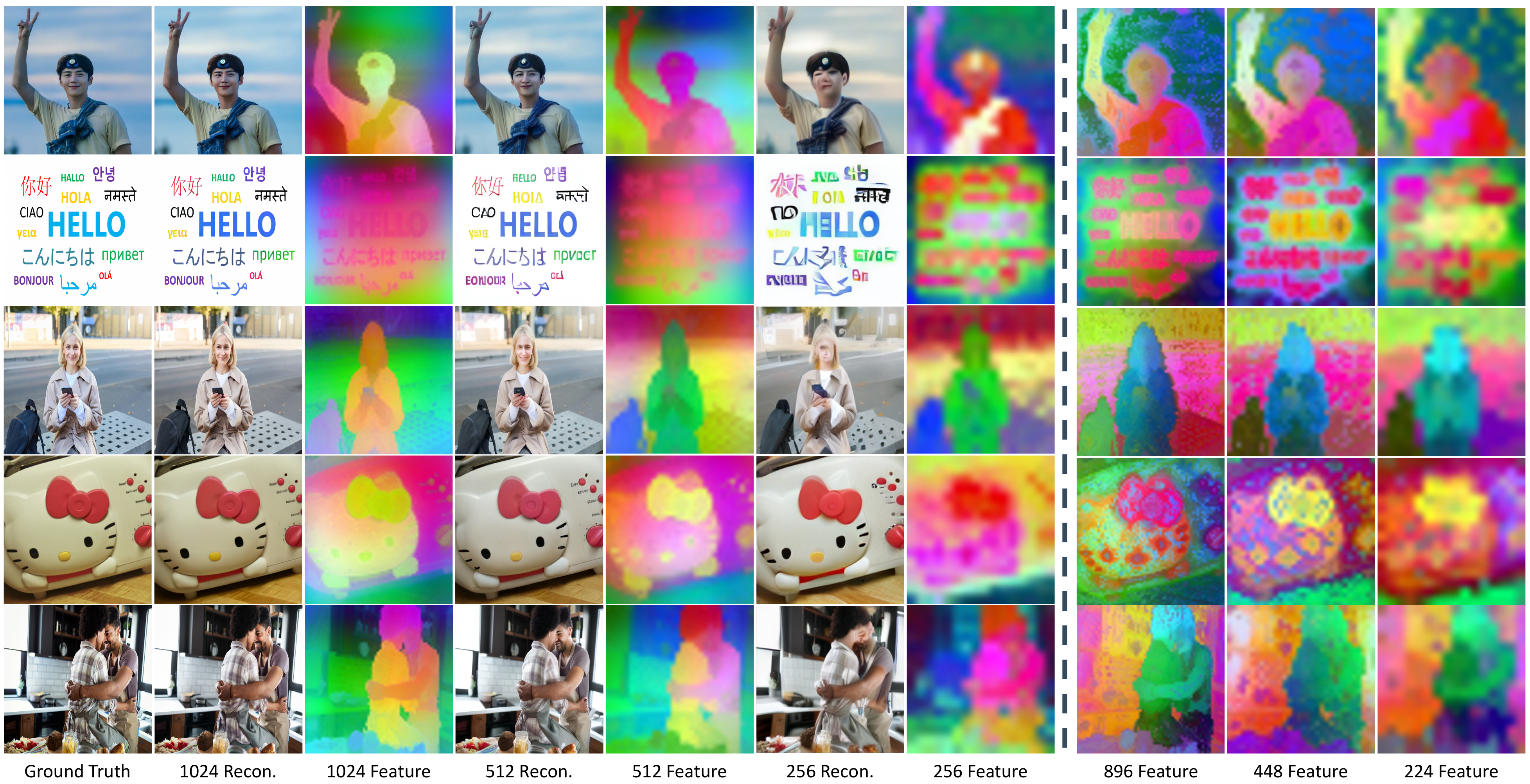}  % 替换为你的图片路径
    % Teaser标题：简洁突出核心贡献，建议加粗
    \caption{\textbf{Visualization of SVG reconstruction.} For each reconstructed image, we also present the PCA visualization of its DINOv3 features, and DINOv2 features are shown in the right three columns. The numbers indicate the resolution of the input images.}
    \label{fig:rec_compare}  % 标签：方便正文引用（如 \ref{fig:teaser}）
    \vspace{-10pt}
\end{figure*}

% Building upon the analysis in~\Cref{subsec:rethinking}, we argue that semantic discriminability in the latent representation space is also critical for generative modeling. A natural approach is to exploit the feature space of self-supervised vision foundation models, such as DINO~\citep{dinov3}, as a substitute. The DINO feature space offers strong semantic discriminability and, in contrast to global semantic–oriented representations such as CLIP~\citep{clip}, preserves rich local visual information. Nevertheless, it remains insufficient for generative tasks, as it discards substantial information related to color and fine-grained perceptual details, as shown in …. To address this limitation, we propose to construct a novel feature space for visual generation that preserves the semantic discriminability of DINO features while augmenting them with residual perceptual detail information to enhance generative capacity.

% Concretely, the~\ours~autoencoder consists of two components. The~\ours~Encoder augments the frozen DINO encoder with a lightweight Residual Encoder built on a Vision Transformer, and the outputs of the two encoders are concatenated along the channel dimension to obtain the complete~\ours~feature. The~\ours~Decoder follows the VAE decoder design from~\citep{ldm}, translating the~\ours~features into pixel-space images.

\noindent\textbf{\ours~DiT.}
We use the Unified Next-DiT~\citep{lumina2} architecture as our backbone, which treats text and image tokens as a joint sequence, enabling natural cross-modal interactions and allowing seamless task expansion. The Unified Next-DiT architecture is a scalable single-stream variant similar to that used in the state-of-the-art open-source VAE-based text-to-image model Z-Image~\citep{z-image}. We adopt this single-stream design to achieve greater parameter efficiency and to jointly process text and DINO features. We train the backbone directly on high-dimensional VFM(DINOv3) feature space, using the flow matching objective defined in~\Cref{equ:fm}. In our framework, we use the \textbf{DINOv3-ViT-S/16+ encoder}, which maps an $H \times W \times 3$ image to a $(H/16) \times (W/16) \times 384$ feature representation.

\noindent\textbf{\ours~Training Pipeline.}
Training proceeds in two stages. In the first stage, we train \textit{autoencoder-P}, \textit{autoencoder-R} separately from scratch. Specifically, \textit{autoencoder-R} is optimized with both reconstruction losses and distribution-matching strategy on its residual branch and decoder following~\citep{svg}. In the second stage, we train \ours~DiT equipped with \textit{autoencoder-P}, following a progressive schedule (see training details).
% During the first training stage, we optimize only the Residual Encoder and the~\ours~Decoder with the reconstruction loss defined in~\citep{ldm}. However, this design introduces a critical problem: because the DINO encoder is frozen, the decoder tends to over-rely on the Residual Encoder, while the outputs of the two encoders exhibit mismatched numerical ranges, thereby undermining the semantic discriminability inherited from DINO. To mitigate these issues and better preserve the semantic structure of the~\ours~Encoder, we introduce two complementary strategies. First, with a fixed probability, we replace the entire output of the Residual Encoder with a mask token, encouraging the decoder to depend more on the frozen DINO features. Second, we explicitly align the Residual Encoder outputs with the DINO feature distribution, ensuring that the additional residual dimensions do not distort the original semantic space. In the second stage, we train~\ours~Diffusion under the settings of SiT~\citep{sit}. For stability, we apply QK-Norm and normalize the~\ours~feature space when training the diffusion model.
\subsection{Scaling SVG to Higher Resolution}
SVG~\citep{svg} and RAE~\citep{rae} primarily focused on learning generative diffusion models within VFM representation spaces under low-resolution settings. In this work, we extend this line of inquiry by examining the behavior and effectiveness of SVG for high-resolution generation. 

We observe distinct resolution-dependent behaviors when reconstructing images from DINOv3 features, as illustrated in~\Cref{fig:rec_compare}. While reconstructions from low-resolution inputs suffer from degradation in fine structures, high-resolution inputs yield substantially more detailed and faithful results. This indicates that DINOv3 representations inherently preserve detailed visual cues effectively at higher resolutions. Crucially, this capability suggests that the DINOv3 encoder alone is relatively sufficient for high-resolution reconstruction, obviating the need for an auxiliary residual encoder. Furthermore, relying exclusively on VFM representations offers a more generalized and reusable paradigm compared to hybrid architectures. Motivated by both the representation sufficiency and the desire for a streamlined, versatile framework, we configure the residual encoder in the original SVG Autoencoder as optional, omitting it during high-resolution reconstruction or generation.

\section{Experiments}
\label{sec:experiments}
In this section, we describe the training recipe of \ours, then validate the feasibility and effectiveness of the proposed \ours~through extensive experiments.

\begin{figure*}[t]  % [t] 表示优先放在页顶；双栏跨栏用 figure*，单栏用 figure
    \centering  % 图片居中对齐
    % 引入图片：width=0.8\textwidth 控制宽度（80%文本宽，避免过满），height可省略（按比例缩放）
    \includegraphics[width=\textwidth]{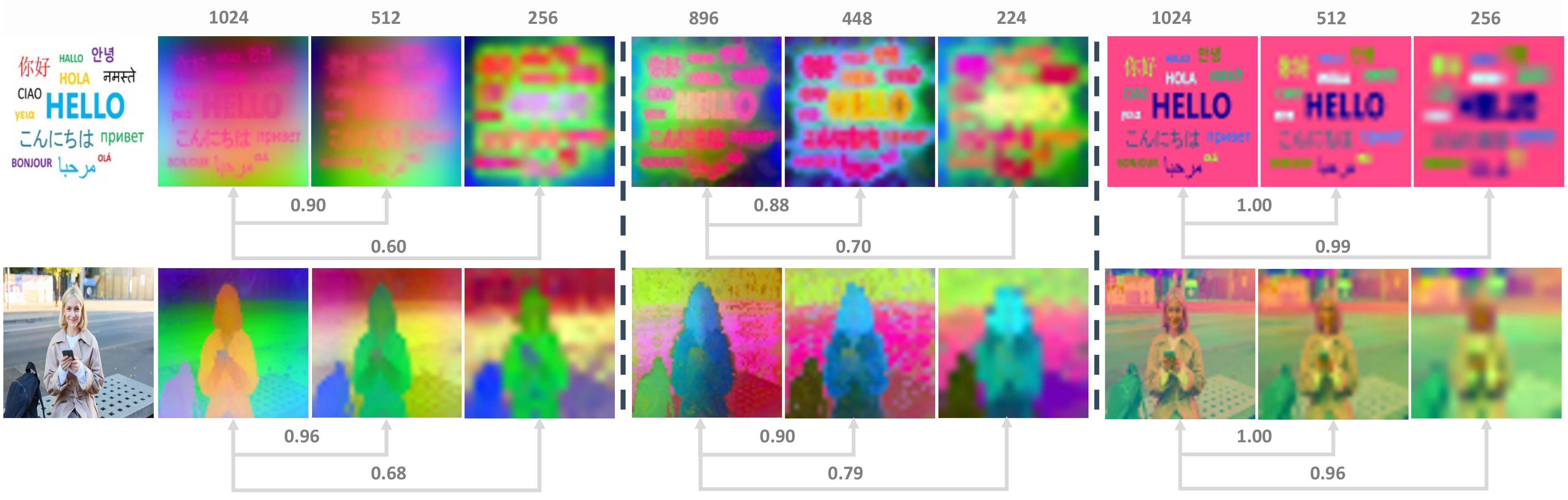}  % 替换为你的图片路径
    % Teaser标题：简洁突出核心贡献，建议加粗
    \caption{\textbf{Comparison of VAE and DINO features.}
    We visualize the PCA projections of features extracted by DINOv3 (Left), DINOv2 (Middle), and the VAE (Right). Cosine similarity across different resolutions is computed by downsampling higher-resolution features to match the spatial size of the lower ones. Overall, VAE features exhibit limited semantic structure, yet demonstrate stronger scale invariance compared to DINO features.
    }
    \label{fig:rec_sim_compare}  % 标签：方便正文引用（如 \ref{fig:teaser}）
\end{figure*}

% Specifically, we investigate the following key questions:

% \begin{itemize}[leftmargin=*, itemsep=2pt, parsep=1pt]
%     \item Can \ours, as a latent text-to-image diffusion model without VAE, achieve competitive generative quality, high training efficiency (\Cref{tab:sys_cmp}), fast inference (\Cref{tab:fewstep}), and favorable scaling properties (\Cref{tab:scaling})?
%     \item Does the \ours~feature space provide task-general representations applicable across diverse vision tasks (\Cref{tab:ablation_all}, \Cref{fig:edit})?
%     % \item Are the choices of VFMs (\Cref{tab:ablation_recon1}) and the components of the SVG Encoder (\Cref{tab:ablation_all}) reasonable?
% \end{itemize}

\subsection{Model Training}
\label{subsec:expsetups}

\noindent\textbf{Training Details of \ours~Autoencoder.}

The autoencoder is trained with a progressive strategy. We first pre-train the model on ImageNet (Data A) for 40 epochs at a fixed resolution of 256×256. Then, during the multi-resolution fine-tuning stage, we continue training using native-resolution images from a 3M-sample dataset (Data B). In this phase, the model is trained at an anchor resolution of 512×512 for 10M seen images, followed by 1024×1024 for an additional 6M seen images.

\begin{table}[t]
\centering
\caption{
\textbf{Overview of the datasets} used across different training stages. 
Cap. Sample Ratio denotes the probability of sampling short, middle, and long captions during training for each dataset, which controls the caption-length distribution seen by the model.}
\label{tab:dataset_overview}
\resizebox{1.0\textwidth}{!}{
\begin{tabular}{lccccc}
\toprule
\textbf{Data} & \textbf{Type} & \textbf{Number} & \textbf{Description} & \textbf{Caption Length} & \textbf{Cap. Sample Ratio} \\
\midrule

% ---- Reconstruction ----
\multirow{2}{*}{Reconstruction}
& A & 1.2M & ImageNet General Data & - & - \\
& B & 3M   & High-quality Realistic Data & - & - \\

\midrule

% ---- Generation ----
\multirow{3}{*}{Generation}
& C & 60M & High-quality General Data & Short, Middle, Long & (0.10, 0.35, 0.55) \\
& D & 15M & High-quality Realistic Data & Short, Middle, Long & (0.10, 0.35, 0.55) \\
& E & 1M  & High-aesthetic Data & Short, Middle, Long & (0.00, 0.00, 1.00) \\

\bottomrule
\end{tabular}
}
\end{table}

\noindent\textbf{Training Details of \ours~DiT.} 

We adopt the Unified Next-DiT architecture from Lumina-Image-2.0~\citep{lumina2} as the backbone of our diffusion transformer. For text conditioning, we utilize the Gemma2-2B large language model, which possesses strong multilingual capabilities, to extract rich textual embeddings. We set the maximum text token length to 256 to balance long-caption modeling capability and training efficiency at first three stages. In high quality data tuning state, the maximum text token length is set to 512. Each image in Data (C, D, E) is annotated with bilingual captions (Chinese and English) in three lengths: short, middle, and long. During training, we adopt a mixed sampling strategy that selects both the caption language and its length. The sampling probabilities for short, middle, and long captions are provided in Table~\ref{tab:dataset_overview}, and the language sampling ratio is fixed to 0.2 for Chinese and 0.8 for English.

\begin{figure*}[t]
    \centering
    \includegraphics[width=\textwidth]{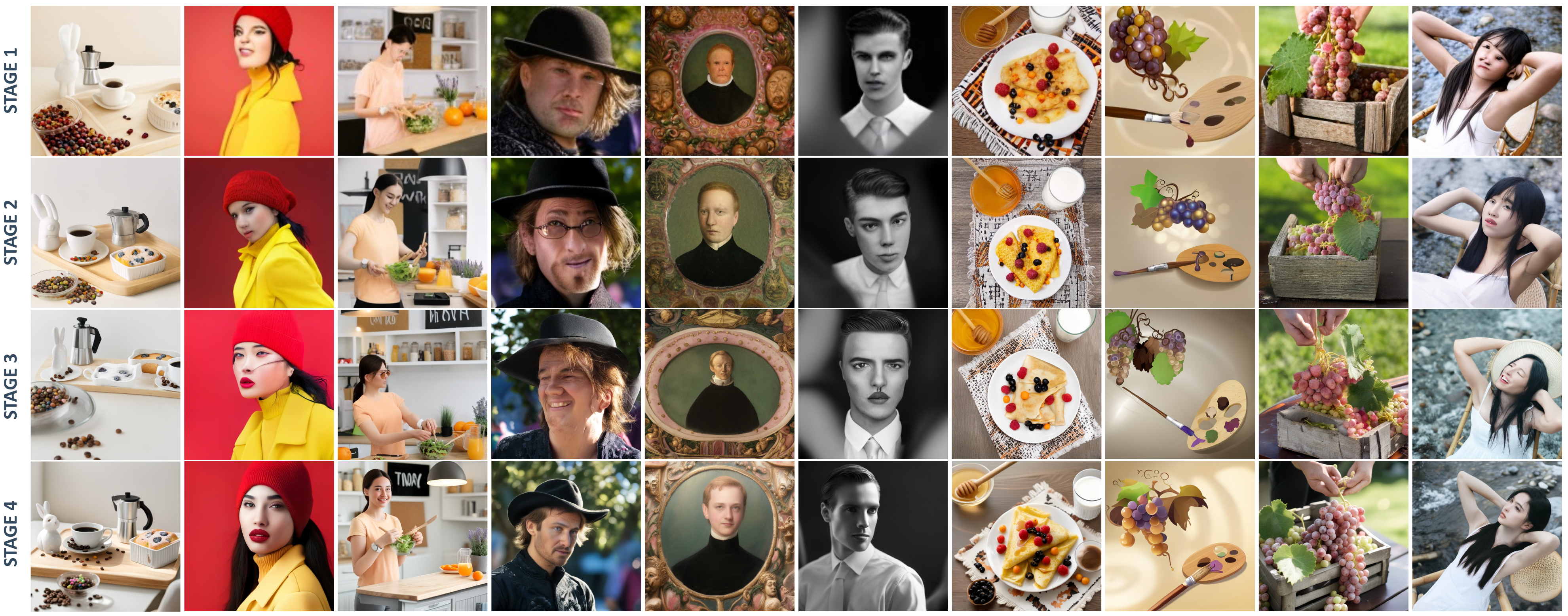}
    \caption{\textbf{Comparison of SVG-T2I four-stage results.} The generated images become progressively more detailed and aesthetically refined as the stages advance.}
    \label{fig:compare_four_stages}
\end{figure*}

We train \ours~which is equipped with autoencoder-P using a multi-stage progressive training strategy. In the first two stage, the model is trained at low resolution and middle resolution on 60M samples (Data C) to establish robust text–image alignment and capture low-frequency structures. In the third stage, we transfer the learned knowledge to higher resolutions, enabling the model to refine fine-grained visual details using 15M samples (Data D). In the final stage, \ours~is fine-tuned on 1M high-quality aesthetic samples (Data E) to further enhance its ability to synthesize realistic and visually appealing outputs. As shown in Figure~\Cref{fig:compare_four_stages}, the visual quality improves steadily across stages.

% After the four-stage training, we obtain \ours-P. However, purely frozen DINO features tend to lose high-frequency details and exhibit mild color cast artifacts, which can negatively affect downstream applications such as image editing. To address this issue while efficiently leveraging the pretrained \ours-P, we further fine-tune \ours-R—equipped with autoencoder-R on the same 1M high-quality samples to learn residual compensatory features. All the three variants demonstrate strong performance in image generation, with \ours-R providing an effective and lightweight solution to mitigate the loss of fine details and color distortions for downstream task such as image editing.

% Finally, we obtain two model variants of SVG-T2I-P/R.

\subsection{Main results}

\noindent\textbf{Evaluation.} We evaluated \ours~through both quantitative and qualitative metrics. We report performance on GenEval~\citep{geneval} and DPG-Bench~\citep{dpgbench} to evaluate all-around capabilities of \ours~following their official protocol. All images used for evaluation were generated at a high resolution of $1024 \times 1024$. Our SVG-T2I model successfully scales the VFM representation paradigm for large-scale T2I generation, achieving competitive performance across these two benchmarks. On GenEval~\cite{geneval} (\Cref{tab:geneval_comparison}), our final model, SVG-T2I, attains an overall score of 0.74, matching the performance of models like SD3-Medium~\citep{sd3} and significantly surpassing SDXL~\citep{sdxl} and DALL-E 2~\citep{dalle3}.  Furthermore, on DPG-Bench (\Cref{tab:dpg_comparion}), SVG-T2I achieves an overall score of 85.78, placing it statistically comparable to top VAE-based diffusion models such as FLUX.1~\citep{flux2024} and HiDream-I1-Full~\citep{hidreami1}.

\begin{table}[ht]
\centering
\caption{Model configuration of the proposed \ours~DiT architecture.}
\resizebox{\textwidth}{!}{
\begin{tabular}{lccccccc}
\toprule
\textbf{Model} & \textbf{Params} & \textbf{Patch Size} & \textbf{Dimension} & \textbf{Head} & \textbf{KV Heads} & \textbf{Layers} & \textbf{Pos. Emb.} \\
\midrule
SVG-T2I & 2.6B & 1 & 2304 & 24 & 8 & 26 & M-RoPE \\
\bottomrule
\end{tabular}
}
\end{table}

\begin{table}[ht]
\centering
\caption{Training configuration of \ours~Decoder across different stages.}
\resizebox{\textwidth}{!}{
\begin{tabular}{lccccccc}
\toprule
\textbf{Stage} & \textbf{Anchor Resolution} & \textbf{\#Images} & \textbf{Training Steps (K)} & \textbf{Batch Size} & \textbf{\#Seen Samples} \\
\midrule
Fixed Low Res. & $256^2$ & 1.2M & 94K & 512 & 48M \\
Multi Middle Res. & $512^2$ & 3M & 78K & 128 & 10M  \\
Multi High Res. & $1024^2$ & 3M & 190K & 32 & 6M  \\

\bottomrule
\end{tabular}
}
\end{table}

\begin{table}[ht]
\centering
\caption{Training configuration of \ours~DiT across different stages.}
\resizebox{\textwidth}{!}{
\begin{tabular}{lccccccc}
\toprule
\textbf{Stage} & \textbf{Anchor Resolution} & \textbf{\#Images} & \textbf{Training Steps (K)} & \textbf{Batch Size} & \textbf{\#Seen Samples}\\
\midrule
Multi Low Res.  & $256^2$ & 60M & 91K & 1536 & 140M  \\
Multi Middle Res.  & $512^2$ & 60M & 90K & 768 & 70M  \\
Multi High Res.  & $1024^2$ & 15M & 44K & 768 & 34M  \\
HQ Tuning.  & $1024^2$ & 1M & 40K & 768 & 30M  \\
\bottomrule
\end{tabular}
}
\end{table}

\begin{table*}[t]
\centering
\caption{Evaluation of text-to-image generation ability on \textbf{GenEval}~\citep{geneval} benchmark. $^\dagger$ refer to the methods using LLM rewriter.}
\label{tab:geneval_comparison}
\resizebox{\textwidth}{!}{
\begin{tabular}{lccccccc}
\toprule
\textbf{Model} & \textbf{Single Obj.} & \textbf{Two Obj.} & \textbf{Counting} & \textbf{Colors} & \textbf{Position} & \textbf{Color Attri.} & \textbf{Overall}$\uparrow$ \\
\midrule
\multicolumn{7}{c}{\textit{Discrete Generation (Autoregressive) }} \\
\midrule
% \multirow{13}{*}{\rotatebox[origin=c]{90}{\textit{Unified}}} 
 Chameleon~\citep{chameleon} & - & - & - & - & - & - & 0.39 \\
 % LWM [42] & 0.93 & 0.41 & 0.46 & 0.79 & 0.09 & 0.15 & 0.47 \\
 % SEED-X [23] & 0.97 & 0.58 & 0.26 & 0.80 & 0.19 & 0.14 & 0.49 \\
 % TokenFlow-XL [59] & 0.95 & 0.60 & 0.41 & 0.81 & 0.16 & 0.24 & 0.55 \\
 % ILLUME [76] & 0.99 & 0.86 & 0.45 & 0.71 & 0.39 & 0.28 & 0.61 \\
 % Janus& 0.97 & 0.68 & 0.30 & 0.84 & 0.46 & 0.42 & 0.61 \\
 % Transfusion & - & - & - & - & - & - & 0.63 \\
 Emu3-Gen$^{\dagger}$~\cite{emu3} & 0.99 & 0.81 & 0.42 & 0.80 & 0.49 & 0.45 & 0.66 \\
 Show-o~\cite{show-o} & 0.98 & 0.80 & 0.66 & 0.84 & 0.31 & 0.50 & 0.68 \\
 Janus-Pro-7B~\citep{janus-pro}  & 0.99 & 0.89 & 0.59 & 0.90 & 0.79 & 0.66 & 0.80 \\
% MetaQuery-XL$^{\dagger}$ [57] & - & - & - & - & - & - & 0.80 \\
\midrule
\multicolumn{7}{c}{\textit{VAE-based Generation (Diffusion) }} \\
\midrule
% \multirow{8}{*}{\rotatebox[origin=c]{90}{\textit{Gen. Only}}} 
 PixArt-$\alpha$~\citep{pixartalpha} & 0.98 & 0.50 & 0.44 & 0.80 & 0.08 & 0.07 & 0.48 \\
 SDv2.1~\citep{ldm} & 0.98 & 0.51 & 0.44 & 0.85 & 0.07 & 0.17 & 0.50 \\
 DALL-E 2~\citep{dalle3} & 0.94 & 0.66 & 0.49 & 0.77 & 0.10 & 0.19 & 0.52 \\
 SDXL~\citep{sdxl} & 0.98 & 0.74 & 0.39 & 0.85 & 0.15 & 0.23 & 0.55 \\
 DALL-E 3~\citep{dalle3} & 0.96 & 0.87 & 0.47 & 0.83 & 0.43 & 0.45 & 0.67 \\
 Lumina-Image-2.0$^{\dagger}$~\citep{lumina2} & - & 0.87 & 0.67 & - & - & 0.62 & 0.73 \\
 SD3-Medium~\citep{sd3} & 0.99 & 0.94 & 0.72 & 0.89 & 0.33 & 0.60 & 0.74 \\
 FLUX.1-dev$^{\dagger}$~\citep{flux2024} & 0.98 & 0.93 & 0.75 & 0.93 & 0.68 & 0.65 & 0.82 \\
\midrule
\multicolumn{7}{c}{\textit{Representation Generation (Diffusion)}} \\
\midrule
% \multirow{13}{*}{\rotatebox[origin=c]{90}{\textit{}}} 
% \textbf{SVG-T2I (Stage4-16K)}$^\dagger$ & 0.97 & 0.91 & 0.49 & 0.84 & 0.63 & 0.60 & 0.74 \\
\textbf{SVG-T2I}$^\dagger$ & 0.94 & 0.89 & 0.49 & 0.89 & 0.69 & 0.62 & 0.75 \\
\bottomrule
\end{tabular}
}
\end{table*}
\begin{table}[ht]
\centering
\caption{Evaluation of text-to-image generation ability on \textbf{DPG}~\citep{dpgbench} benchmark.}
\resizebox{0.95\textwidth}{!}{
\begin{tabular}{l c c c c c c}
\toprule
Model & Global & Entity & Attribute & Relation & Other & Overall $\uparrow$ \\
\midrule
\multicolumn{7}{c}{\textit{Discrete Generation (Autoregressive)}} \\
\midrule
Janus~\citep{janus} & 82.33 & 87.38 & 87.70 & 85.46 & 86.41 & 79.68 \\
Emu3-Gen~\citep{emu3} & 85.21 & 86.68 & 86.84 & 90.22 & 83.15 & 80.60 \\
Janus-Pro-1B~\citep{janus-pro} & 87.58 & 88.63 & 88.17 & 88.98 & 88.30 & 82.63 \\
Janus-Pro-7B~\citep{janus-pro} & 86.90 & 88.90 & 89.40 & 89.32 & 89.48 & 84.19 \\
\midrule
\multicolumn{7}{c}{\textit{VAE-based Generation (Diffusion)}} \\
\midrule
SD1.5~\citep{ldm} & 74.63 & 74.23 & 75.39 & 73.49 & 67.81 & 63.18 \\
PixArt-$\alpha$~\citep{chen2023pixart_alpha} & 74.97 & 79.32 & 78.60 & 82.57 & 76.96 & 71.11 \\
Lumina-Next~\citep{lumina-next} & 82.82 & 88.65 & 86.44 & 80.53 & 81.82 & 74.63 \\
SDXL~\citep{sdxl} & 83.27 & 82.43 & 80.91 & 86.76 & 80.41 & 74.65 \\
Hunyuan-DiT~\citep{hunyuandit} & 84.59 & 80.59 & 88.01 & 74.36 & 86.41 & 78.87 \\
PixArt-$\Sigma$~\citep{chen2024pixart_sigma} & 86.89 & 82.89 & 88.94 & 86.59 & 87.68 & 80.54 \\
DALL-E 3~\citep{dalle3} & 90.97 & 89.61 & 88.39 & 90.58 & 89.83 & 83.50 \\
FLUX.1 [Dev]~\citep{flux2024} & 74.35 & 90.00 & 88.96 & 90.87 & 88.33 & 83.84 \\
SD3 Medium~\citep{sd3} & 87.90 & 91.01 & 88.83 & 80.70 & 88.68 & 84.08 \\
HiDream-I1-Full~\citep{hidreami1} & 76.44 & 90.22 & 89.48 & 93.74 & 91.83 & 85.89 \\
Lumina-Image 2.0~\citep{lumina2} & - & 91.97 & 90.20 & 94.85 & - & 87.20 \\
\midrule
\multicolumn{7}{c}{\textit{Representation Generation (Diffusion)}} \\
\midrule
% CFG 8 SHIFT10 STEPS25
% \textbf{SVG-T2I (Stage4-32K)} & 88.50 & 91.00 & 91.86 & 92.21 & 91.86 & 85.78 \\
\textbf{SVG-T2I} & 88.50 & 91.00 & 91.86 & 92.21 & 91.86 & 85.78 \\
\bottomrule
\end{tabular}
}
\label{tab:dpg_comparion}

\end{table}

\begin{figure}[t]
    \centering
    \includegraphics[width=\linewidth]{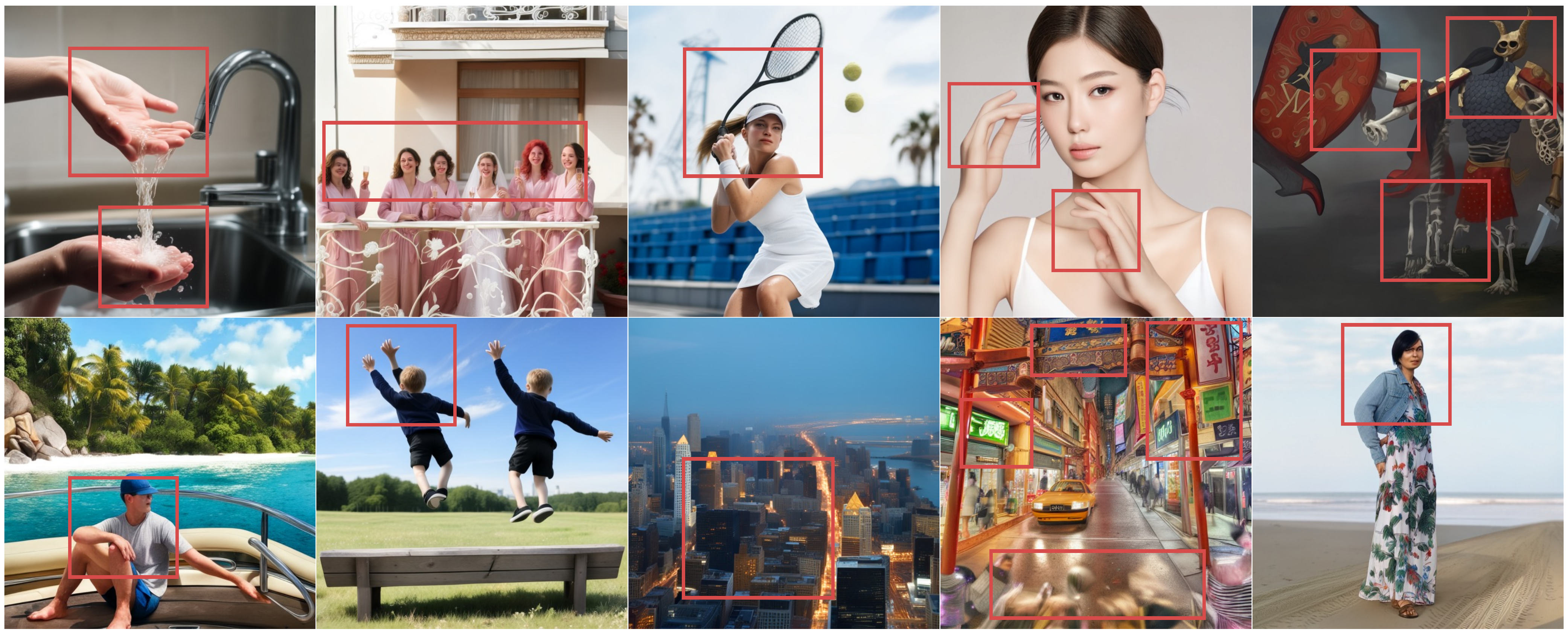}
    \caption{\textbf{Failure cases of SVG-T2I.}
    These examples illustrate that SVG-T2I may struggle with generating highly detailed human faces, accurate finger structures, and reliable text rendering, which typically require more specialized data and larger training compute.}
    \label{fig:failure_case}
\end{figure}

\subsection{Analysis}
% Existing self-supervised learning methods are capable of producing features that are both semantically rich and sufficiently informative for accurate reconstruction. In principle, these representations encode enough fine-grained details to support high-quality generation. However, when training involves multiple resolutions, especially low-resolution inputs, the feature consistency across scales is not guaranteed. Introducing explicit constraints to enforce similarity of features across different resolutions could enhance the reconstruction potential of low-resolution representations, allowing SSL-derived features to retain both semantically-rich and fine-grained visual information across scales.

\noindent\textbf{Limitations of current VFM features.} Existing self-supervised learning methods produce representations that capture both high-level semantic context and fine-grained visual detail, offering a strong basis for downstream reconstruction and generation. In principle, these representations are largely self-sufficient. However, this self-sufficiency is critically challenged when the training paradigm involves multiple input resolutions. As represented in~\Cref{fig:rec_sim_compare}, VAE features exhibit nearly resolution-invariant behavior. Their cross-resolution consine similarity is close to 1.0, whereas DINOv3 and DINOv2 features vary more substantially. This observation indicates that VFM-derived features undergo non-negligible shifts across scales.

When a VFM encoder uses a fixed patch or receptive-field size (e.g., 16×16) across inputs of different absolute resolution, the semantic granularity and effective compression ratio of each patch vary systematically with scale: a patch on a low-resolution image aggregates a much larger portion of the scene, producing strongly compressed, detail-poor features; the identical patch size on a high-resolution image captures finer, predominantly local texture and structural detail. Because VFM encoders are typically optimized to produce semantically discriminative tokens rather than to preserve uniform local detail, they are particularly sensitive to this scale-dependent shift in the semantic/texture balance. By contrast, reconstruction-oriented encoders (e.g., VAEs) do not explicitly account for the semantic content present in each encoded region; instead, they primarily aim to capture sufficient local information for pixel-level reconstruction, leading to a more uniform and resolution-stable allocation of representation capacity.

Accordingly, for semantic visual encoders used for diffusion modeling, maintaining stable cross-resolution behavior emerges as an important optimization goal. The training pipeline may need to incorporate mechanisms that encourage consistent feature geometry and help preserve the fidelity of fine-grained details across scales.

\noindent\textbf{Limitations of SVG-T2I.}
While SVG-T2I demonstrates strong generation capability across diverse scenarios, several limitations remain. As shown in~\Cref{fig:failure_case}, the model occasionally struggles to produce highly detailed human faces, particularly in regions requiring fine-grained spatial consistency, such as eyes, eyebrows. Similarly, the generation of anatomically accurate fingers continues to be challenging, a common failure mode in generative models, often resulting in distorted shapes or incorrect topology when pose complexity increases. SVG-T2I also exhibits limited reliability in text rendering. These shortcomings largely stem from the insufficient coverage of such fine-grained cases in the training corpus, as well as the substantial computational demand required to model high-frequency patterns and precise geometric relationships. Addressing these limitations will likely require more specialized datasets and additional training compute.

\section{Conclusion}
In this work, we successfully extended the original SVG framework to large-scale, high-resolution text-to-image synthesis, culminating in the SVG-T2I model. Our work validates the feasibility of training a high-quality T2I model from scratch based on VFM representations, achieving generative metrics comparable to modern advanced methods and demonstrating the potential of the VFM semantic space as an effective latent manifold for high-resolution synthesis. To foster further research and ensure reproducibility, we have fully open-sourced the training, inference, and evaluation code, along with the model weights, hoping to benefit the academic community. However, in the course of this research, we also identified a critical challenge: existing VFM encoders (such as DINOv2~\citep{dinov2} and DINOv3~\citep{dinov3}) produce representations with poor internal consistency when encoding the same image at different input resolutions. This resolution-dependent feature instability directly compromises the T2I model's ability to generalize across various sizes and maintain generation quality, underscoring the necessity for future research to focus on scale-invariance. Overall, we believe that the strategic use and refinement of a powerful VFM latent space, as demonstrated in this work, present a highly promising avenue toward achieving a truly unified representation for diverse visual tasks.

\bibliography{main}
\bibliographystyle{iclr2026_conference}

\clearpage
\appendix

\begin{table*}[ht]
\centering
\caption{Hyperparameter setup of \ours~Autoencoder.}
\label{tab:hyperparam}
\renewcommand{\arraystretch}{0.8}
\resizebox{0.55\textwidth}{!}{%
\begin{tabular}{lcc}
\toprule
 & Autoencoder-P & Autoencoder-R \\
\midrule
\textbf{Encoder} & & \\
Base model      & DINOv3-s16p & DINOv3-s16p \\
Downsample Ratio      & $16 \times 16$ & $16 \times 16$ \\
Latent dim. & 384 & 392 \\
Residual branch       & /   & ViT-S-RoPE \\
Training mode         & Frozen & Frozen \\
Params                & 29M   & 51M \\
\midrule
\textbf{Decoder} & & \\
Channels dim. & \multicolumn{2}{c}{[512, 256, 256, 128, 128]} \\
Out channels           & 3   & 3 \\
Z channels             & 384  & 392 \\
Params                & 43M   & 43M \\
\midrule
\textbf{Optimization} & & \\
Optimizer             & Adam & Adam \\
lr                    & 1e-4 & 1e-4 \\
$(\beta_1,\beta_2)$   & (0.5, 0.9) & (0.5, 0.9) \\
\bottomrule
\end{tabular}%
}
\end{table*}

\begin{table*}[ht]
\centering
\caption{Hyperparameter setup of \ours~DiT.}
% \label{tab:hyperparam}
\renewcommand{\arraystretch}{0.8}
\resizebox{0.55\textwidth}{!}{%
\begin{tabular}{lc>{\columncolor{gray!10}}c}
\toprule
 & SVG-T2I-L & Lumina-Image-2.0 \\
\midrule
\textbf{Architecture} & & \\
Downsample Ratio      & $16 \times 16$ & $8\times 8$ \\
Latent dim.           & \textbf{384} & \textbf{16} \\
Num. layers           & 26   & 26 \\
Hidden dim.           & 2304 & 2304 \\
Num. heads            & 24   & 24 \\
Params                & 2.6B & 2.6B \\
Base-encoder          & Autoencoder-P & Flux-VAE \\
\midrule
\textbf{Optimization} & & \\
Optimizer             & AdamW & AdamW \\
lr                    & 2e-4 & 2e-4 \\
$(\beta_1,\beta_2)$   & (0.9, 0.95) & (0.9, 0.95) \\
\midrule
\textbf{Interpolants} & & \\
$\alpha_t$            & $1-t$ & $1-t$ \\
$\sigma_t$            & $t$   & $t$ \\
Training objective     & v-prediction & v-prediction \\
Sampler               & Euler & Euler \\
\bottomrule
\end{tabular}%
}
\end{table*}

\section{More Qualitative Results}
We provide additional qualitative results of SVG-T2I with $1080 \times 1080$ resolution. 
These results further demonstrate the diversity and visual quality of the proposed approach.

\clearpage
\begin{figure}[t]
    \centering
    % === 第一行 ===
    \begin{subfigure}{\linewidth}
        \centering
        \includegraphics[width=\linewidth]{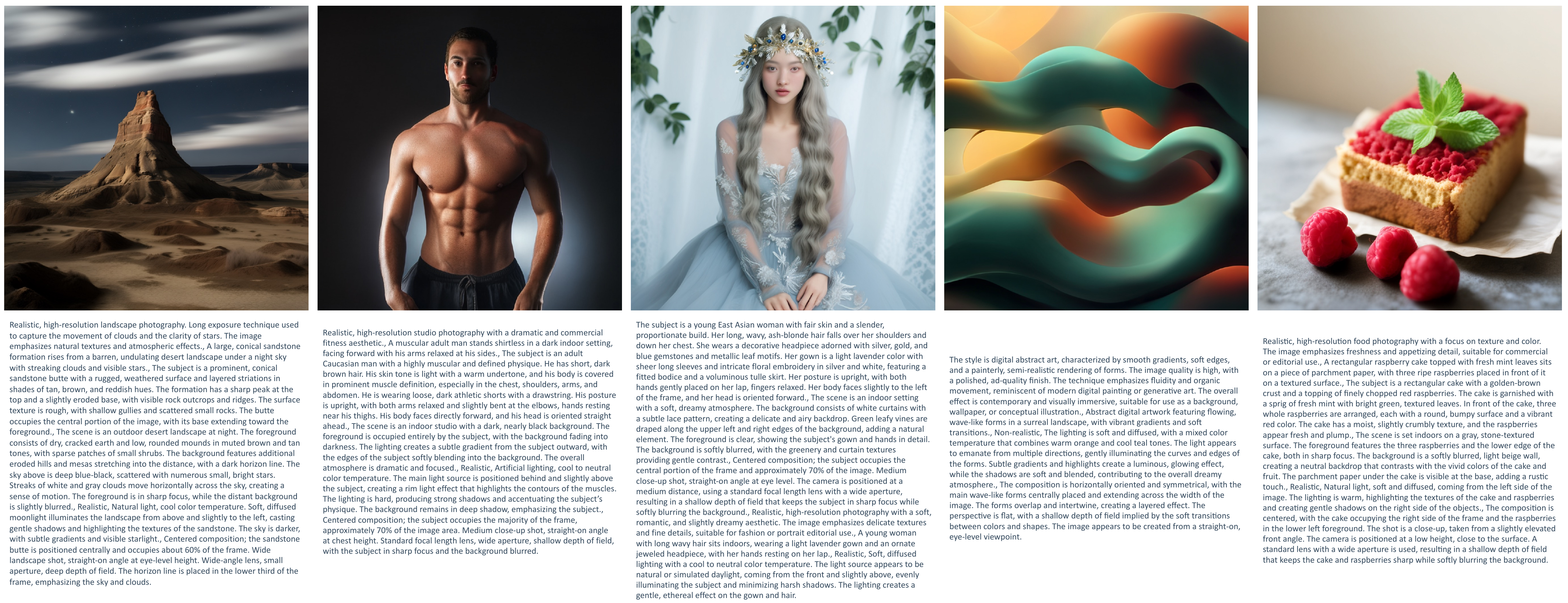}
    \end{subfigure}

    % === 第二行 ===
    \begin{subfigure}{\linewidth}
        \centering
        \includegraphics[width=\linewidth]{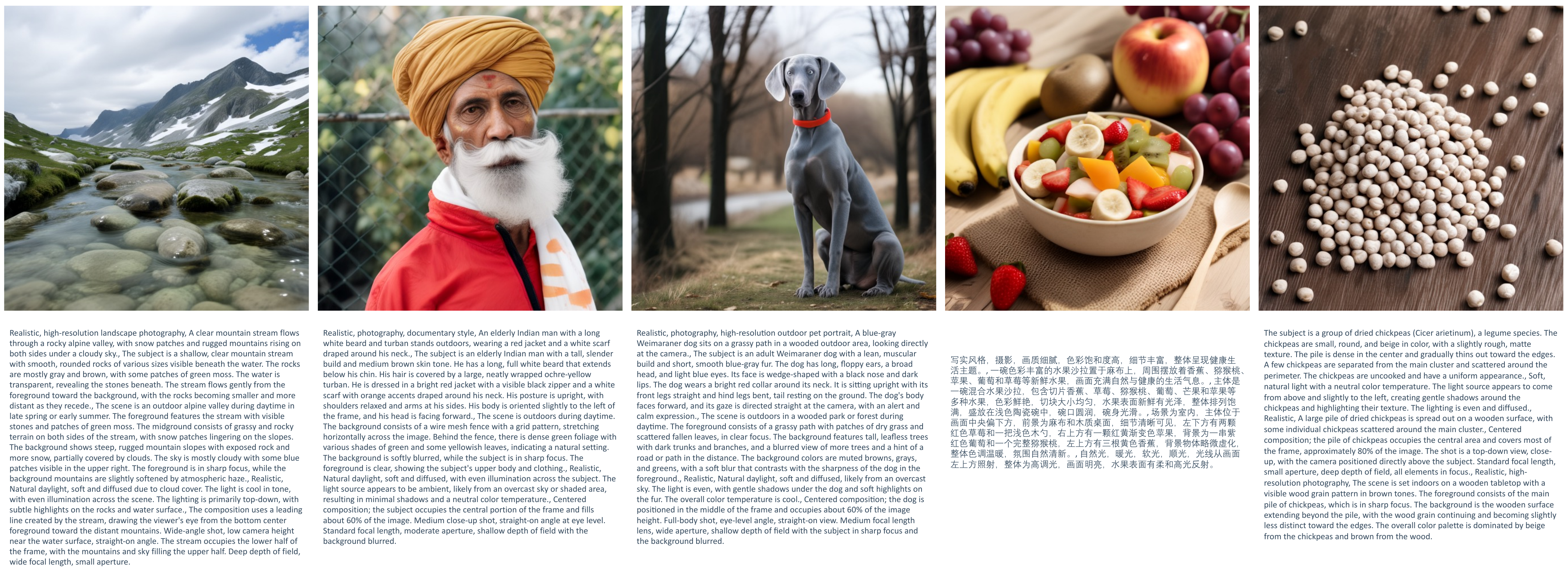}
    \end{subfigure}
    
    % === 第三行 ===
    \begin{subfigure}{\linewidth}
        \centering
        \includegraphics[width=\linewidth]{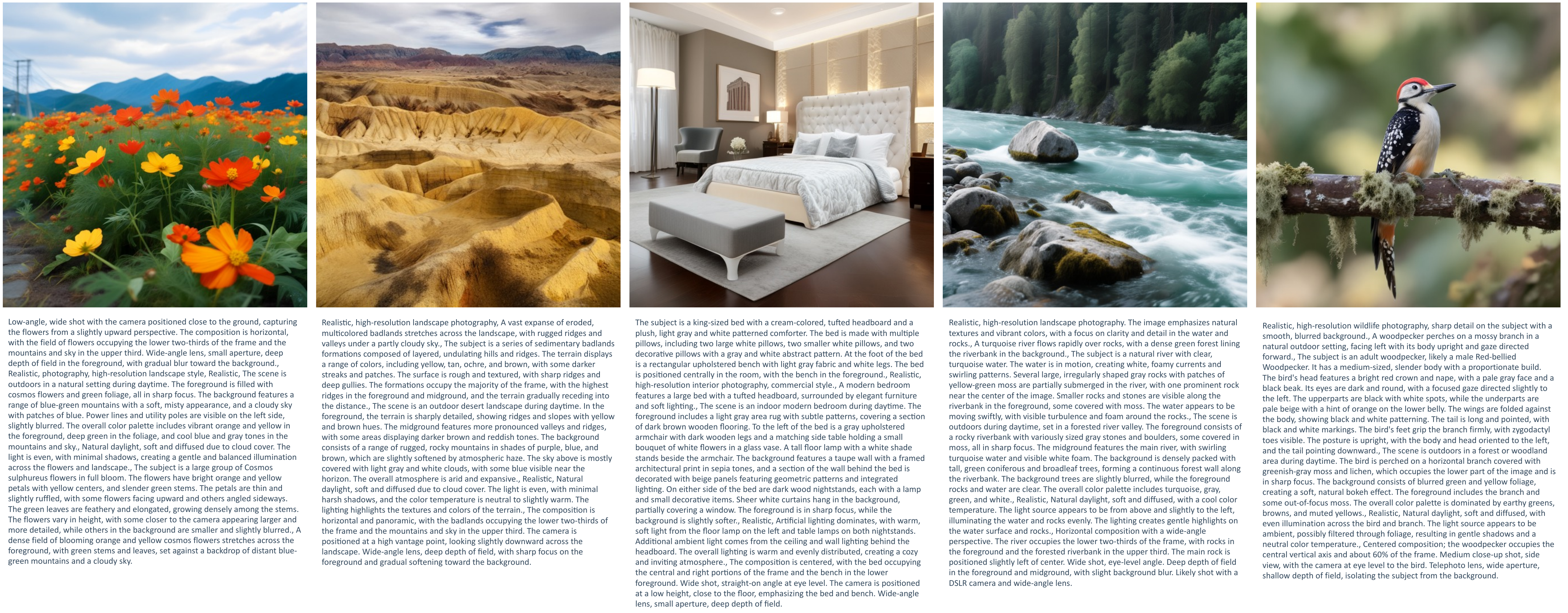}
    \end{subfigure}

    % === 第四行 ===
    \begin{subfigure}{\linewidth}
        \centering
        \includegraphics[width=\linewidth]{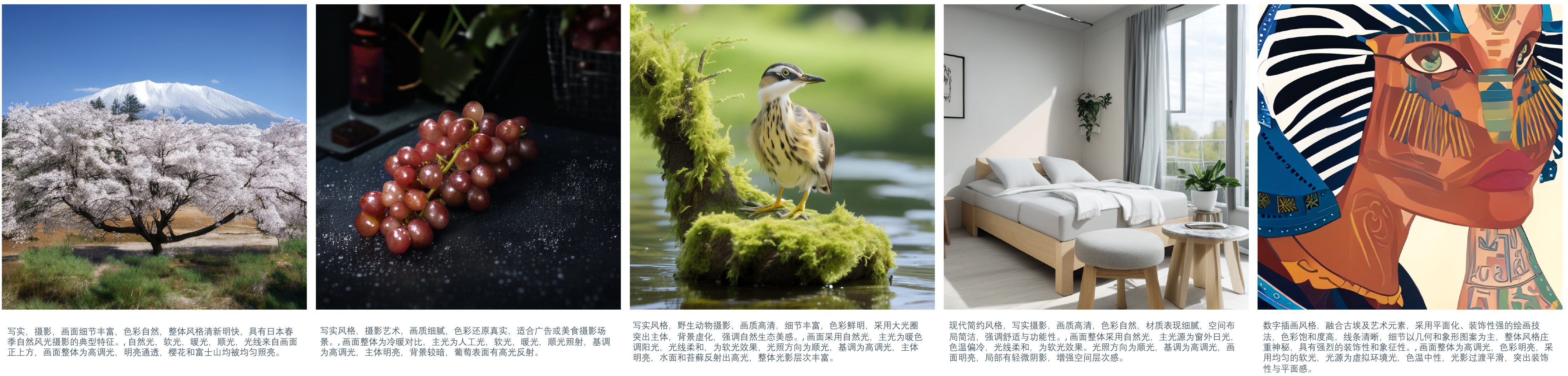}
    \end{subfigure}

    \caption{\textbf{Visualization of SVG-T2I at 1080×1080 resolution.} 
    These examples highlight the model’s ability to produce high-quality, text-aligned visual content with fine-grained details and strong spatial consistency.}
    \label{fig:feature_viz_rabbit}
\end{figure}
\clearpage

\section{Prompts used in~\Cref{fig:showcases}}
\label{app:prompt}

Here we summarize the prompts/instructions used in \Cref{fig:showcases}, which can be directly input into the corresponding model to reproduce our generation results.

Column \#1:

\begin{itemize}
    \item Case \#1:
    
    Realistic, high-resolution photography with a commercial still-life style. The image emphasizes clarity, texture, and color accuracy, suitable for food and beverage advertising or editorial use., Two clear wine glasses filled with white wine are placed on a rustic wooden table, surrounded by green grapes, bread, wine corks, and a bottle of white wine in the background., The main subjects are two transparent wine glasses, each filled with pale yellow-white wine. The glasses are made of clear glass with long stems and round bowls. The wine is clear and slightly golden, with visible reflections on the glass surfaces. The glasses are positioned close together, with the left glass slightly in front of the right. Both glasses are upright and stable on the table. In the foreground, there are clusters of green grapes with smooth, round shapes and translucent skin. Several loose grapes are scattered near the base of the glasses. To the right of the glasses, there is a piece of rustic brown bread with a rough texture and visible crust. In the background, a wire basket contains multiple cylindrical beige wine corks, and a glass bottle of white wine is partially visible., The scene is set indoors on a rustic wooden table with a natural, weathered texture and light brown color. The foreground features green grapes in sharp focus, with some grapes and corks scattered around the base of the glasses. The midground contains the wine glasses, bread, and a wire basket of corks. The background is slightly blurred and includes a glass bottle of white wine and a wooden wall with a warm, neutral tone. The overall color palette consists of natural wood, green, beige, and pale yellow tones, creating a cozy and inviting atmosphere., Realistic, The lighting is natural and soft, with a warm color temperature. The main light source comes from the left side, casting gentle highlights on the glass surfaces and creating soft reflections. The lighting is even, with subtle shadows under the glasses and grapes, contributing to a bright and inviting tone., The composition is centered, with the two wine glasses occupying the central area of the frame and taking up about 50\% of the image. The shot is a close-up, taken at a slightly high angle, focusing on the glasses and surrounding elements. The camera is positioned at table height, angled slightly downward. A standard lens with a wide aperture is used, resulting in a shallow depth of field that keeps the foreground and main subjects sharp while softly blurring the background.

    \item Case \#2:
    
    The subject is a young East Asian woman with fair skin and a slender, proportionate build. Her long, wavy, ash-blonde hair falls over her shoulders and down her chest. She wears a decorative headpiece adorned with silver, gold, and blue gemstones and metallic leaf motifs. Her gown is a light lavender color with sheer long sleeves and intricate floral embroidery in silver and white, featuring a fitted bodice and a voluminous tulle skirt. Her posture is upright, with both hands gently placed on her lap, fingers relaxed. Her body faces slightly to the left of the frame, and her head is oriented forward., The scene is an indoor setting with a soft, dreamy atmosphere. The background consists of white curtains with a subtle lace pattern, creating a delicate and airy backdrop. Green leafy vines are draped along the upper left and right edges of the background, adding a natural element. The foreground is clear, showing the subject's gown and hands in detail. The background is softly blurred, with the greenery and curtain textures providing gentle contrast., Centered composition; the subject occupies the central portion of the frame and approximately 70\% of the image. Medium close-up shot, straight-on angle at eye level. The camera is positioned at a medium distance, using a standard focal length lens with a wide aperture, resulting in a shallow depth of field that keeps the subject in sharp focus while softly blurring the background., Realistic, high-resolution photography with a soft, romantic, and slightly dreamy aesthetic. The image emphasizes delicate textures and fine details, suitable for fashion or portrait editorial use., A young woman with long wavy hair sits indoors, wearing a light lavender gown and an ornate jeweled headpiece, with her hands resting on her lap., Realistic, Soft, diffused lighting with a cool to neutral color temperature. The light source appears to be natural or simulated daylight, coming from the front and slightly above, evenly illuminating the subject and minimizing harsh shadows. The lighting creates a gentle, ethereal effect on the gown and hair.

    \item Case \#3: 
    
    Realistic, high-resolution landscape photography, A snow-covered mountain range rises behind a calm lake, with the peaks and surrounding forest reflected clearly in the water under a bright blue sky., The subject is a group of rugged mountain peaks covered in white snow, with sharp ridges and rocky outcrops. The mountains are flanked by steep, forested slopes with dark green coniferous trees. The lower slopes and the area around the lake are dusted with snow. The reflection of the mountains and trees is visible in the still, pale greenish water of the lake, forming a near-symmetrical image., The scene is an outdoor alpine landscape during daytime in winter. The foreground consists of the shallow, clear lake water with a sandy and slightly muddy bottom, reflecting the mountains and sky. The middle ground features the lake’s edge lined with snow and a dense band of dark green coniferous trees. The background is dominated by the snow-capped mountain range, with rocky faces and patches of snow, set against a cloudless, vivid blue sky. The foreground water is clear and detailed, while the background mountains are sharp and well-defined., Realistic, Natural sunlight, cool color temperature, hard light. The light source is from above and slightly to the left, illuminating the snow and casting subtle shadows on the mountain slopes. The lighting is even and high-key, enhancing the clarity of the snow and the reflection in the water., Symmetrical composition with the mountain range centered in the frame and its reflection forming a vertical axis. Wide landscape shot, straight-on angle at eye level. The mountains occupy the upper half of the frame, while the lake and its reflection fill the lower half. Wide-angle lens, small aperture, deep depth of field.

    \item Case \#4: 
    
    photorealistic, The image features three cupcakes adorned with colorful sprinkles and encased in checkered wrappers. The cupcakes are positioned on a pink surface with scattered sprinkles around them. In the background, there is a small green leaf, possibly a mint leaf, adding a hint of freshness to the composition., The foreground of the image showcases three cupcakes. Each cupcake is generously topped with a colorful assortment of sprinkles that include shades of pink, white, red, green, and blue. The cupcakes are wrapped in a checkered brown and black paper, providing a neat and structured appearance. The upper surface of the cupcakes is golden brown, indicating they are well-baked, with a soft, crumbly texture visible. They are compactly arranged, filling the bottom half of the image., The background of the image consists of a smooth pink surface, which enhances the vibrant colors of the sprinkles. Among the scattered sprinkles, there is a small green leaf, likely a mint leaf, offering a contrast to the otherwise pink and colorful visuals. The setting suggests a modern and bright environment, focused on accentuating the cupcakes., solid color, close-up, standard lens, natural light, central composition, frontal view

    \item Case \#5: 
    
    realistic photography, frontal view, blurred, central composition, A vibrant red rose is prominently featured in the foreground, with droplets of water on its petals. The background shows green foliage and bushes, likely indicating a garden setting. The image captures the delicacy and color of the rose against a blurred backdrop of greenery., natural light, close-up, macro lens.

    \item Case \#6: 
    
    realistic, high-resolution photography, A Bengal cat with green eyes lies on a brown surface, facing the camera with its head slightly tilted., The subject is an adult Bengal cat. It has a muscular build and short, dense fur with a golden-brown base and dark brown rosette and stripe patterns. The cat's face is broad with a pinkish-brown nose, white muzzle, and long white whiskers. Its ears are upright and pointed, with pinkish inner fur. The cat's eyes are large, round, and bright green, with vertical slit pupils. Its front legs are tucked under its body, and its head is slightly tilted to the left. The cat's gaze is directed straight at the camera, and its expression appears calm and alert., The scene is indoors. The foreground consists of a brown fabric surface, likely a couch or cushion, which is in sharp focus. The background is a smooth, dark brown wall, softly blurred. The overall color palette is warm, dominated by shades of brown and gold., Realistic, Natural light, soft and diffused, coming from the front left. The lighting is warm, evenly illuminating the cat's face and fur, with gentle shadows under the chin and on the right side of the face., Centered composition; the cat's face occupies the central area of the frame, filling about 70\% of the image. Medium close-up shot, eye-level angle. The camera is positioned close to the subject, using a standard or short telephoto lens. Shallow depth of field, with the cat in sharp focus and the background blurred.
    
\end{itemize}

Column \#2:

\begin{itemize}
    \item Case \#1:
    
    Realistic, high-resolution photography with a classic portrait style. The image emphasizes texture and detail in clothing and accessories, with a focus on elegance and sophistication., A woman with curly brown hair is dressed in a fur-collared coat and a black hat with ornate embroidery, standing against a dark, softly lit background., The subject is a woman, likely a young adult, with light skin and voluminous, curly brown hair that frames her head and falls around her ears. She wears a black hat adorned with intricate gold and silver embroidery featuring floral and leaf patterns. Her ears are visible, and she wears dangling, multi-stone earrings. She is dressed in a coat with a thick, grayish fur collar that frames her neck and shoulders. Her posture is upright, with her head and body facing directly forward., The scene is indoors with a dark, softly blurred background. The background features deep brown and reddish hues, creating a warm and subdued atmosphere. The foreground is occupied by the subject’s upper body and clothing, which are in sharp focus. The background is out of focus, providing a sense of depth and isolating the subject., Realistic, Soft, warm lighting, likely artificial, coming from the front and slightly above the subject. The light gently illuminates the subject’s hair, hat, and fur collar, creating subtle highlights and soft shadows. The background remains darker, enhancing the subject’s prominence., Centered composition; the subject occupies the central portion of the frame and fills most of the vertical space. Medium close-up shot, straight-on angle at eye level. The camera is positioned close to the subject, using a standard or short telephoto lens. Shallow depth of field, with the subject in sharp focus and the background blurred.

    \item Case \#2:
    
    Realistic style, photography, finely detailed, high color saturation, rich in details, overall conveying a healthy lifestyle theme. A bowl of colorful fruit salad placed on a burlap cloth, surrounded by fresh fruits such as bananas, kiwis, apples, grapes, and strawberries, the scene full of natural and healthy life vibes. The main subject is a bowl of mixed fruit salad, including sliced bananas, strawberries, kiwis, grapes, mangoes, and apples, with vivid colors, evenly cut pieces, fresh and glossy fruit surfaces, arranged generously, served in a light-colored ceramic bowl with a rounded rim and smooth body. The scene is indoors, with the main subject positioned slightly lower in the center of the frame; the foreground features burlap and a wooden tabletop with clear details. In the lower-left corner, there are two red strawberries and a light wooden spoon; in the upper-right corner, a red-yellow gradient apple; the background contains a bunch of reddish-purple grapes and a whole kiwi, with three yellow bananas in the upper-left corner. Background objects are slightly blurred, the overall color tone is warm, and the atmosphere is natural and fresh. Natural warm soft light, front lighting from the upper-left side, overall high-key lighting, bright scene, with soft highlights reflecting off the fruit surfaces.

    \item Case \#3:
    
    A vibrant watercolor botanical illustration featuring red and yellow tulip flowers, depicted in a close-up pattern style with abstract blue and green backgrounds., The image is a high-quality, semi-realistic botanical illustration created in watercolor. The style combines realistic floral forms with expressive, abstract background elements. The technique features wet-on-wet blending, visible brushstrokes, and splatter effects, characteristic of contemporary botanical art. The overall effect is lively, decorative, and artistic, suitable for use in textiles, wallpapers, or stationery., The main subjects are several tulip flowers, shown in various stages of bloom. The tulips have elongated, slightly curved petals with a mix of red, yellow, and orange hues, blending softly at the edges. The petals are rendered with visible watercolor gradients and subtle textural variations, giving a sense of translucency and freshness. The green stems and leaves are painted with loose, expressive brushstrokes, adding to the lively aesthetic. The flowers are depicted in a natural, upright orientation, with some petals overlapping and others angled differently, creating a dynamic arrangement., The lighting is implied through the use of watercolor techniques, with soft, diffused highlights and gentle shading that suggest natural daylight. The color temperature is warm, with the red and yellow petals contrasting against the cooler blue and green background. The overall tone is high key, with bright, luminous colors and minimal shadow, enhancing the fresh and uplifting mood., The environment is an abstract, artistic background composed of overlapping washes of blue, green, and yellow watercolor, with splashes and drips that evoke a sense of spontaneity and movement. The background is mostly clear, with the floral elements standing out against the colorful, painterly backdrop. The overall atmosphere is bright, cheerful, and energetic, reminiscent of a spring or early summer setting., Non-realistic, The composition is a close-up, pattern-like arrangement with a repeating motif. The tulip flowers are distributed diagonally and vertically across the frame, filling most of the image space. The perspective is flat and frontal, with the flowers overlapping and intersecting, creating a sense of depth through layering. The image uses a centered and balanced composition, with the flowers occupying the majority of the visual field. The depth of field is shallow, with all elements rendered in sharp focus due to the illustrative nature.

    \item Case \#4:
    
    The scene is indoors with a dark, neutral background. The foreground features soft, out-of-focus shapes in the lower left and right corners, possibly fabric or shadows, which frame the subject. The background is uniformly dark, providing contrast to the subject's lighter clothing and hair. The overall atmosphere is subdued and formal., Centered composition; the subject occupies the central portion of the frame, filling approximately 60\% of the image. Medium close-up shot, straight-on angle at eye level. The camera is positioned close to the subject. Shallow depth of field, with the foreground and background softly blurred, focusing attention on the subject's upper body and hair., Realistic, black-and-white photography with a formal portrait style. The image has a high-resolution, professional finish, emphasizing texture and contrast., Soft, diffused lighting with a cool tone. The light source appears to come from the front and slightly above, illuminating the subject's hair and shirt evenly. The lighting creates gentle shadows and a smooth gradient across the background., Realistic, The subject is a young Caucasian man with fair skin and a proportionate build. His hair is short, dark, and neatly combed to the side, with a smooth texture. He is wearing a white collared shirt and a light-colored tie, suggesting formal attire. His posture is upright, with shoulders squared and body facing forward. The subject's head is oriented straight ahead., A young Caucasian man with neatly styled hair is pictured in a formal setting, wearing a collared shirt and tie, with his upper body visible against a dark background.

\end{itemize}

Column \#3:

\begin{itemize}
    \item Case \#1:
    
    Realistic, high-resolution wildlife photography, An adult Bengal tiger lies on the ground in a grassy outdoor setting, facing the camera with its body partially visible and its gaze directed forward., The subject is an adult Bengal tiger. It has a large, muscular build with orange fur and prominent black stripes running along its body and face. The tiger's face is broad with a pink nose, white fur around the mouth and chin, and white whiskers. Its ears are rounded with black backs and white inner fur. The tiger's eyes are yellow-green, and it is gazing directly at the camera. Its body is reclined on the ground, with the front legs extended forward and the head held upright. The tiger's expression is calm and alert., The scene is outdoors during daytime. The foreground contains a pile of pale beige straw or dried grass, which is in sharp focus. Behind the tiger, there is a patch of green grass and a horizontal log with a rough brown texture. The background consists of more green grass and blurred beige ground, creating a natural habitat atmosphere. The foreground is clear, while the background is softly blurred., Natural daylight, soft and diffused, with even illumination across the tiger's face and body. The light source appears to be from above and slightly to the front, producing gentle shadows and highlighting the tiger's fur texture., Centered composition; the tiger's head and upper body occupy the central area of the frame, filling about 60\% of the image. Medium close-up shot, eye-level angle. The camera is positioned at the tiger's eye height, straight-on. Medium focal length, moderate aperture, shallow depth of field., Realistic.

    \item Case \#2:
    
    The scene is outdoors in a garden or natural setting during daytime. The foreground features the sharply focused cosmos flowers and a few green stems and buds. The background is blurred, displaying additional purple cosmos flowers and green foliage, with circular bokeh highlights in shades of purple, green, and yellow. The overall color palette includes purples, greens, and warm yellow tones, creating a vibrant and lively atmosphere., Realistic, photography, high-resolution macro style with a focus on natural detail and color. The image emphasizes clarity and vibrancy, typical of botanical or nature photography., Realistic, Several blooming purple cosmos flowers are clustered together outdoors, with sunlight illuminating their petals and casting soft shadows., The subjects are cosmos flowers (Cosmos bipinnatus) in full bloom. The flowers have broad, delicate petals in a soft purple hue with subtle gradients and slightly ruffled edges. The centers are bright yellow-orange, surrounded by a ring of small, dark-tipped stamens. The petals are thin and semi-translucent, catching the light. The flowers are at various angles, with the central flower facing forward and slightly upward, while others are angled to the left or right. The stems are slender and green, with some unopened buds visible below the flowers., Centered composition with the main flower group occupying the middle of the frame. Medium close-up shot, straight-on angle at flower height. The main flowers fill about 60\% of the frame. Shallow depth of field achieved with a wide aperture, rendering the background and some foreground elements softly blurred. Likely taken with a DSLR camera and a macro or standard lens., Natural sunlight, warm color temperature, soft light quality. The light source comes from the upper left, creating gentle highlights on the petals and subtle shadows on the lower right sides of the flowers. The lighting emphasizes the translucency and texture of the petals.

    \item Case \#3:
    The image features a close-up of two glass cups filled with Dalgona coffee. The foreground shows a cup with whipped coffee foam on top of a milk layer. The background includes a blurred cup of the same drink and a stone-like surface with scattered coffee beans., realistic, blurred, close-up, natural light, frontal view, standard lens, central composition.

\end{itemize}

\end{document}